\documentclass[utf8]{FrontiersinHarvard}
\usepackage{url,hyperref,lineno,microtype,subcaption}
\usepackage[onehalfspacing]{setspace}
\usepackage{multirow}
\usepackage{booktabs} 

\def\keyFont{\fontsize{8}{11}\helveticabold }
\def\firstAuthorLast{Sanderson and Matuszewski}
\def\Authors{Edward Sanderson\,$^{*,1}$ and Bogdan J. Matuszewski\,$^{1}$}
\def\Address{$^{1}$Computer Vision and Machine Learning (CVML) Group, School of Engineering, University of Central Lancashire, Preston, UK}
\def\corrAuthor{University of Central Lancashire, Fylde Road, Preston, UK, PR1 8HE}

\def\corrEmail{esanderson4@uclan.ac.uk}

\begin{document}
\onecolumn
\firstpage{1}

\title[Polyp Segmentation Generalisability of Pretrained Backbones]{Polyp Segmentation Generalisability of Pretrained Backbones} 

\author[\firstAuthorLast ]{\Authors} 
\address{} 
\correspondance{} 

\extraAuth{}

\maketitle

\vspace{-0.8cm}
{\tiny
 \keyFont{Keywords: polyp, segmentation, generalisability, self-supervised, pretraining, colonoscopy}}

\section{Introduction}

Due to the relatively low amount of annotated data available for training polyp segmentation models, e.g. \cite{fcbformer}, which typically take the form of an autoencoder with UNet-style skip connections \citep{unet}, pretraining of the encoder, also known as the backbone, is typically performed. Until very recently, this was almost exclusively done in a supervised manner with ImageNet-1k \citep{imagenet}. However, we recently demonstrated that pretraining backbones in a self-supervised manner generally provides better fine-tuned performance \citep{ssl4gieme}, and that models with ViT-B \citep{vit} backbones typically perform better than models with ResNet50 \citep{resnet} backbones.

In this paper, we extend this recent work to consider generalisability. I.e., we assess the performance of models on a different dataset to that used for fine-tuning, accounting for variation in network architecture and pretraining pipeline (algorithm and dataset). This reveals how well models with different pretrained backbones generalise to data of a somewhat different distribution to the training data, which will likely arise in deployment due to different cameras and demographics of patients, amongst other factors.

\section{Analysis}

We consider 12 polyp segmentation models pretrained and fine-tuned in a previous study \citep{ssl4gieme}, specifically those fine-tuned on Kvasir-SEG \citep{kvasirseg}. Each model is either a ResNet50 encoder with a DeepLabV3+ \citep{deeplab} decoder, or a ViT-B encoder with a DPT \citep{dpt} decoder. Additionally, each model was pretrained on either Hyperkvasir-unlabelled \citep{hyperkvasir} or ImageNet-1k in a self-supervised manner using either MoCo v3 \citep{mocov3}, Barlow Twins \citep{barlow} (ResNet50 only), or MAE \citep{mae} (ViT-B only), or pretrained in a supervised manner (ImageNet-1k only), or not pretrained at all.

We evaluate performance on the full CVC-ClinicDB dataset \citep{cvc} with mDice, mIoU, mPrecision, and mRecall. The results are reported in Table \ref{tab:kvasirtocvc}, where we also specify each model's rank on each metric, as well as any change in rank relative to the model's evaluation on the Kvasir-SEG test set \citep{ssl4gieme}. The results show that self-supervised pretraining on ImageNet-1k generally provides the best generalisation, that supervised pretraining on ImageNet-1k is generally better than self-supervised pretraining on Hyperkvasir-unlabelled, and that any considered pretraining is better than no pretraining. These findings are consistent with the evaluation on the Kvasir-SEG test set.

\begin{table}[!t]
\caption{\label{tab:kvasirtocvc}Performance of models fine-tuned on the Kvasir-SEG training set and tested on CVC-ClinicDB. In addition to reporting the value of each metric, we also indicate the rank of each model, as well as any change in this rank relative to the model's evaluation on the Kvasir-SEG test set. For conciseness, we abbreviate Hyperkvasir-unlabelled to HK, ImageNet-1k to IN, and Barlow Twins to BT.}
\centering
\begin{tabular}{ccccccccccc}
\toprule
Backbone            & \multicolumn{2}{c}{Pretraining} & \multicolumn{2}{c}{mDice} & \multicolumn{2}{c}{mIoU} & \multicolumn{2}{c}{mPrecision} & \multicolumn{2}{c}{mRecall} \\   \cmidrule{2-11}
arch. & Data & Algo. & Value & Rank & Value & Rank & Value & Rank & Value & Rank\\ \hline
\multirow{6}{*}{ResNet50} & \multirow{2}{*}{HK} & MoCo v3           & 0.789 &	9 {\tiny \textcolor{blue}{($\uparrow$1)}} &	0.686 &	10 &	0.785 &	10 &	0.856 &	3 {\tiny \textcolor{blue}{($\uparrow$7)}}

                             \\
                          &                                       & BT      & 0.801 &	8 {\tiny \textcolor{blue}{($\uparrow$1)}} &	0.709 &	8 {\tiny \textcolor{blue}{($\uparrow$1)}} &	0.831 &	 8 {\tiny \textcolor{blue}{($\uparrow$1)}} &	0.848 &	7 {\tiny \textcolor{red}{($\downarrow$2)}}

                                                        \\ \cmidrule{2-11} 
                          & \multirow{3}{*}{IN}          & MoCo v3           &  0.843 &	 1 {\tiny \textcolor{blue}{($\uparrow$3)}} &	0.760 &	1 {\tiny \textcolor{blue}{($\uparrow$3)}} &	0.867 &	3 {\tiny \textcolor{blue}{($\uparrow$5)}} &	0.874 &	1
                                                   \\ 
                          &                                       & BT      & 0.826 &	5 &	0.735 &	6 {\tiny \textcolor{blue}{($\uparrow$1)}} &	0.858 &	6 &	0.854 &	4 {\tiny \textcolor{blue}{($\uparrow$2)}}

                                                        \\ \cmidrule{3-11}
                          &                                       & Supervised        & 0.822 &	6 &	0.735 &	5 &	0.899 &	1 {\tiny \textcolor{blue}{($\uparrow$4)}} &	0.811 &	10 {\tiny \textcolor{red}{($\downarrow$2)}}

                                                        \\ \cmidrule{2-11} 
                          & None                                  & None              & 0.520 &	12 &	0.394 &	12 &	0.496 &	12 &	0.724 &	12

                                                        \\ \cmidrule{1-11}
\multirow{6}{*}{ViT-B}    & \multirow{2}{*}{HK} & MoCo v3           & 0.789 &	10 {\tiny \textcolor{red}{($\downarrow$2)}} &	0.696 &	9 {\tiny \textcolor{red}{($\downarrow$1)}} &	0.812 &	9 {\tiny \textcolor{red}{($\downarrow$2)}} &	0.848 &	8 {\tiny \textcolor{red}{($\downarrow$1)}}

                                                        \\
                          &                                       & MAE               & 0.828 &	3 &	0.743 &	3 &	0.852 &	7 {\tiny \textcolor{red}{($\downarrow$4)}} &	0.858 &	2 {\tiny \textcolor{blue}{($\uparrow$1)}}

 \\ \cmidrule{2-11} 
                          & \multirow{3}{*}{IN}          & MoCo v3           & 0.830 &	2 &	0.742 &	4 {\tiny \textcolor{red}{($\downarrow$2)}} &	0.861 &	4 {\tiny \textcolor{red}{($\downarrow$2)}} &	0.849 &	5 {\tiny \textcolor{red}{($\downarrow$3)}}

                                                        \\
                          &                                       & MAE               & 0.827 &	4 {\tiny \textcolor{red}{($\downarrow$3)}} &	0.746 &	2 {\tiny \textcolor{red}{($\downarrow$1)}} &	0.868 &	2 {\tiny \textcolor{red}{($\downarrow$1)}} &	0.848 &	6 {\tiny \textcolor{red}{($\downarrow$2)}}

                                                        \\ \cmidrule{3-11}
                          &                                       & Supervised        & 0.809 &	7 &	0.717 &	7 {\tiny \textcolor{red}{($\downarrow$1)}} &	0.860 &	5 {\tiny \textcolor{red}{($\downarrow$1)}} &	0.832 &	9

                                                        \\ \cmidrule{2-11} 
                          & None                                  & None              & 0.637 &	11 &	0.519 &	11 &	0.670 &	11 &	0.759 &	11

                                                        \\ \bottomrule
\end{tabular}
\end{table}

\begin{figure}[htp]
\centering
\includegraphics[width=0.9\textwidth]{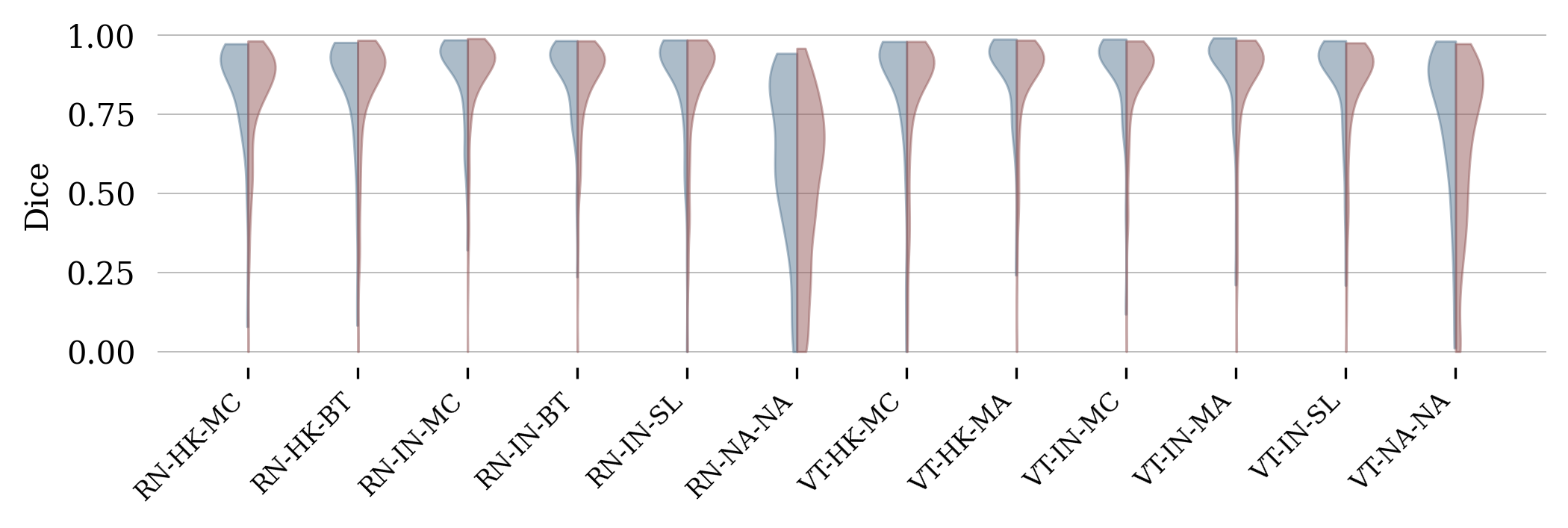}
\caption{Comparison of the distribution of instance-wise Dice score from each model's evaluation on the Kvasir-SEG test set (blue) against the distribution from its evaluation on CVC-ClinicDB (red). For conciseness, we denote ResNet50s with RN, ViT-Bs with VT, Hyperkvasir-unlabelled with HK, ImageNet-1k with IN, MoCo v3 with MC, Barlow Twins with BT, MAE with MA, supervised pretraining with SL, and no pretraining with NA-NA.}
\label{fig:dist_comp}
\end{figure}

However, the model pretrained with MAE on ImageNet-1k, which performs best on the Kvasir-SEG test set, reduces its rank on every metric, notably dropping from rank 1 to 4 on mDice. In contrast, models with a ResNet50 backbone generally improve their ranking, implying greater generalisability than models with a ViT-B backbone, which generally experience a drop in ranking, and the best generalisation is achieved by the model with a ResNet50 backbone that was pretrained on ImageNet-1k using MoCo v3, notably improving from rank 4 to 1 on mDice. To better understand this, we compare the distribution of instance-wise Dice scores from each model's evaluation on the Kvasir-SEG test set against the distribution from its evaluation on CVC-ClinicDB in Fig. \ref{fig:dist_comp}. This indicates that all models experience a drop in overall performance that primarily arises from a higher variance. However, the portion of each distribution for the highest Dice scores shows that most models with ResNet50 backbones achieve better performance on some instances of CVC-ClinicDB than any in the Kvasir-SEG test set, while models with ViT-B backbones fail to exceed their maximum Dice score across the Kvasir-SEG test set when evaluated on CVC-ClinicDB. We verify that all models experience a drop in performance, and quantify the relative drop, in Fig. \ref{fig:dist_comp}, which reveals that most models with ResNet50 backbones do indeed experience less of a drop, potentially as a result of their improvement in maximum Dice score, explaining the improvement in ranking.

\section{Conclusion}
In this paper, we showed that the findings of previous work, regarding pretraining pipelines for polyp segmentation, hold true when considering generalisability. However, our results imply that models with ResNet50 backbones typically generalise better, despite being outperformed by models with ViT-B backbones in evaluation on the test set from the same dataset used for fine-tuning. We expect that this is a result of the larger complexity of the models with ViT-B backbones allowing for overfitting on the distribution underlying the training data. However, this challenges the assumption that the considered pretraining pipelines should help prevent this, and more work is required to better understand the relationships between architecture, pretraining pipeline, and performance on different distributions of data.

\begin{figure}[htp]
\centering
\includegraphics[width=0.75\textwidth]{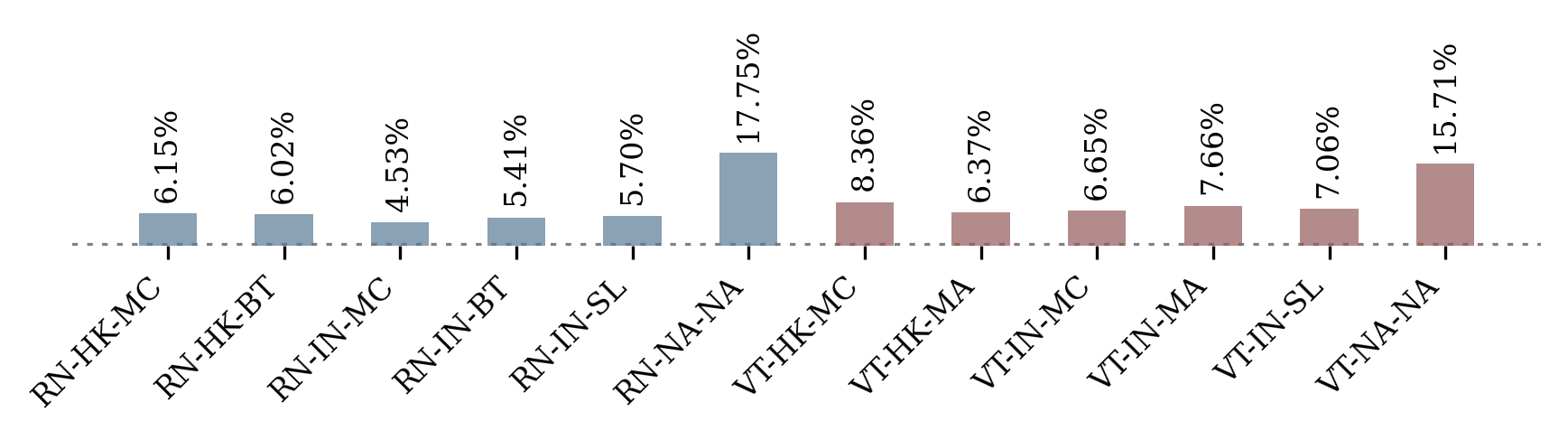}
\caption{Relative drop in mDice from each model's evaluation on the Kvasir-SEG test set to its evaluation on CVC-ClinicDB. For conciseness, we denote ResNet50s with RN, ViT-Bs with VT, Hyperkvasir-unlabelled with HK, ImageNet-1k with IN, MoCo v3 with MC, Barlow Twins with BT, MAE with MA, supervised pretraining with SL, and no pretraining with NA-NA. For clarity, the results for ResNet50 models are coloured blue and the results for ViT-B models are coloured red.}
\label{fig:perf_drop}
\end{figure}

\section*{Conflict of Interest Statement}

The authors declare that the research was conducted in the absence of any commercial or financial relationships that could be construed as a potential conflict of interest.

\section*{Author Contributions}

ES: Conceptualization, Formal analysis, Visualization, Writing – original draft, Writing – review \& editing; BJM: Funding acquisition, Project administration, Resources, Supervision, Writing – review \& editing.

\section*{Funding}
This work was supported by the Science and Technology Facilities Council [grant number ST/S005404/1].

\section*{Data Availability Statement}
This publication is supported by Kvasir-SEG and CVC-ClinicDB which are openly available as cited in the 'References' section of this paper.

\bibliographystyle{Frontiers-Harvard}
\bibliography{test}

\end{document}